# Contrastive ESA: Human Evaluation of Multiple Translations at Once

**Vilém Zouhar**
ETH Zurich

**Roman Grundkiewicz**
Microsoft

**Sara Rajaee**
University of Amsterdam

**Parker Riley**
Google

**Martin Popel**
Charles University

**Rachel Bawden**
Inria

**Philipp Koehn**
JHU

**Marine Carpuat**
University of Maryland

**Tom Kocmi**
Cohere

## Abstract

Current human evaluation of machine translation typically assesses single outputs in isolation, a paradigm that suffers from high annotator noise and cost. We introduce Contrastive Error Span Annotation (cESA), a protocol that presents multiple translations of the source input (text, video, audio, image). In cESA, the annotator sees multiple translations of the same document, marks major and minor error spans, and then assigns a score from 0% to 100% on absolute scale. By allowing annotators to access the shared context across multiple outputs, cESA facilitates more consistent and efficient judgments. We validate cESA using a large-scale human evaluation of English→Japanese translations of 12 models, demonstrating reductions in annotation time and noise compared to standard pointwise evaluation. Unlike existing contrastive ranking methods, cESA yields absolute quality judgments that enable simple, interpretable non-parametric model rankings without the need for post-hoc corrections.[1]

ESA (current)

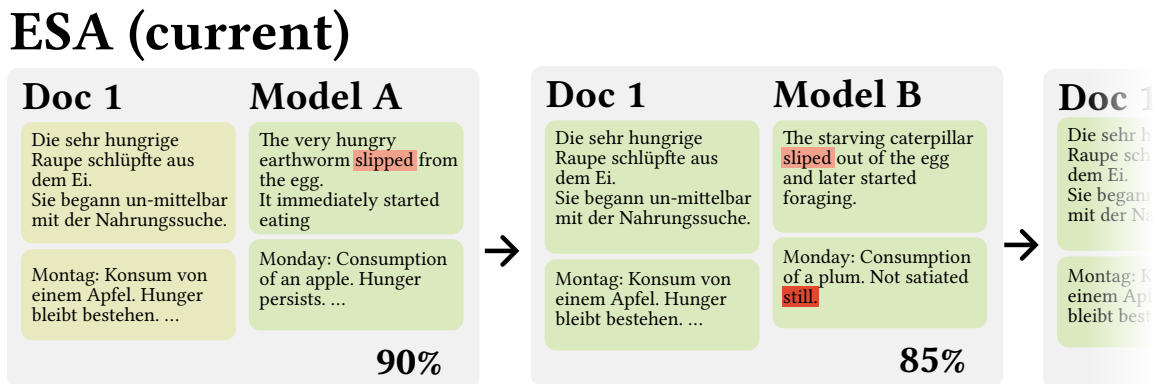

contrastive ESA (proposed)

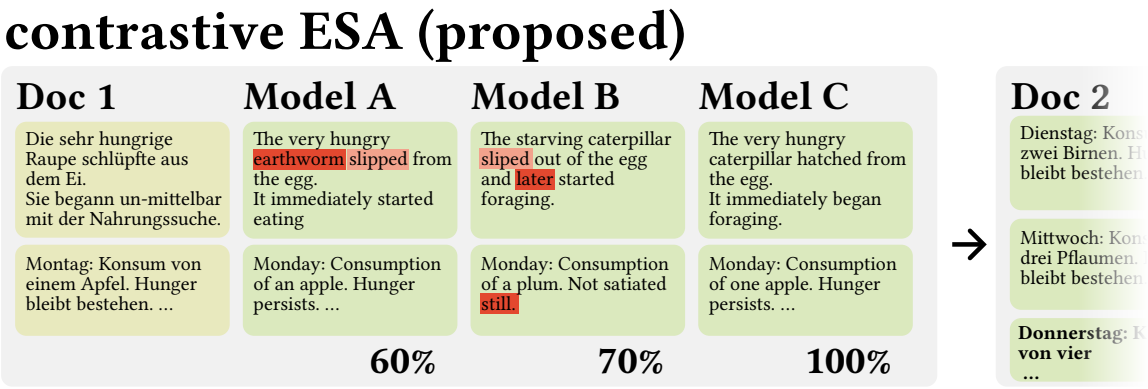

Figure 1: Comparison of pointwise and contrastive ESA. In the standard pointwise ESA (top), some errors go undetected. In contrastive ESA (bottom), multiple outputs are shown at the same time, which also speeds up the annotation.

## 1 Introduction

Human evaluation ultimately decides research direction and deployment decisions. Unfortunately, it comes at a high cost because the annotation noise needs to be balanced by large scale. Many attempts have been made to improve human evaluation of translation by modifying the instructions and types of assessments (Graham et al., 2013a, 2013b, 2017; Freitag et al., 2021a; Kocmi et al., 2024; Zouhar et al., 2025; Ferstler et al., 2026). However, most of these modifications happen within the framework where a source input is shown next to a single translation. In contrast, some approaches show multiple outputs at the same time, but either in a scenario where *all* model outputs can be shown next to each other (Bojar et al., 2016; Novikova et al., 2018) or with only a *subset* of model outputs shown next to each other with post-hoc estimation of model quality (Callison-Burch et al., 2007; Sakaguchi et al., 2014; Sakaguchi and Van Durme, 2018). Showing multiple outputs at the same time is beneficial for multiple reasons, as supported by psychometrics, known as *joint vs. separate evaluation* (Hsee, 1996): (1) finding mistakes in one translation helps in finding mistakes in another translation of the same input (Meyer and Schvaneveldt, 1971), (2) seeing the space of possible translations helps in objective assessment (Hsee, 1996), and (3) sharing the same document context and not duplicating reading times makes the overall annotation faster (Sweller, 1988).

[1] Code & data: github.com/zouharvi/cESA
Try interactive annotated cESA.

In this work, focused on human annotations, we propose simple but effective modifications to the standard pointwise Error Span Annotation (Kocmi et al., 2024) by annotating multiple document-level translation outputs next to each other in the scenario with many outputs in the competition, which cannot all be shown at the same time. The result is a new standardized and tested annotation protocol, cESA (Contrastive Error Span Annotation). We further answer practical questions to make this annotation protocol production-ready at scale:

- What is the optimal number of outputs to show next to each other, what are the tradeoffs, and other necessary user interface aspects?
- What is the nature and severity of biases when showing multiple outputs at the same time?

We define the cESA annotation protocol in Section 3 together with measures of what quantitatively makes a good annotation protocol. We use cESA for human evaluation experimentally in Section 4 on WMT25 translation evaluation (Kocmi et al., 2025b). We show that against standard pointwise ESA, our cESA with three models at the same time can save up to 31% of the annotation cost (average time 77.8s → 53.7s) per item while increasing the annotation quality, as measured by higher annotation stability and inter-annotator agreement.

**Using cESA.** Based on our findings, we encourage future evaluators of translations to adopt this evaluation protocol, which we make available in an existing annotation interface, that can be launched with minimal setup (Listing 1).

## 2 Related Work

The general WMT translation shared task (Kocmi et al., 2025b, inter alia) is an example of a periodic large-scale evaluation effort. Annually, across many language directions, many models are evaluated, which necessitates high-quality yet economic human annotation protocols. For good scientific practices, the annotation protocol itself needs to be formalized and tested before being used as a measurement instrument (Schuff et al., 2023; Thomson et al., 2024; Belz et al., 2024), even if similar approaches might have been ad-hoc used for evaluation without the goal of verifying the quality of the annotation protocol (Popel et al., 2020, inter alia).

```
cat << EOF > campaign.json
{
"info": {
  "assignment": "single-stream",
  "protocol": "cESA"
},
"campaign_id": "my_campaign",
"data": [[
  {
    "src": "Die sehr hungrige Raupe schlüpfte...",
    "tgt": {
      "modelA": "The famished larva hatched...",
      "modelB": "The hungry caterpillar hatched...",
      "modelC": "The very hungry caterpillar ate..."
    }
  }
]]
}
EOF

pip install pearmut
pearmut add campaign.json
pearmut run
```

Listing 1: Minimal bash commands to run cESA evaluation of translation in Pearmut (Zouhar and Kocmi, 2026). We use Pearmut version 1.1.5; documentation.

**Pointwise evaluation.** We distinguish between *pointwise* and *contrastive* evaluation paradigms. Machine translation is commonly evaluated with protocols where a single source and single output is shown. The simplest is Direct Assessment (DA, Graham et al., 2017) with a single-number judgment per translation. More complex, Multidimensional Quality Metrics (MQM, Freitag et al., 2021b) have expert annotators mark erroneous spans, their severity, and their category. The Error Span Annotation (ESA, Kocmi et al., 2024) attempts to strike the balance between depth and economic feasibility by skipping the error categorization and still including the final numerical judgment. Importantly, Knowles and Lo (2025) call for context-aware human evaluation, which improves the calibration of scores and reduces noise.

**Contrastive evaluation.** In contrast, evaluating with absolute scores in isolation is noisy (Knowles and Lo, 2025)[2] and evaluating with ranking or comparison holds promise for more accuracy (Thurstone, 1927; Briakou et al., 2023). Even WMT has used ranking with multiple outputs at the same time (Bojar et al., 2016). This specific evaluation approach showed all model outputs at the same time, which is difficult to scale to document-level evaluation. Recently, Song et al. (2025) extended standard MQM with side-by-side

[2]This problem is exacerbated by common annotation tools not supporting navigating to previous screens to update annotations once, for example, a new error mode is discovered.

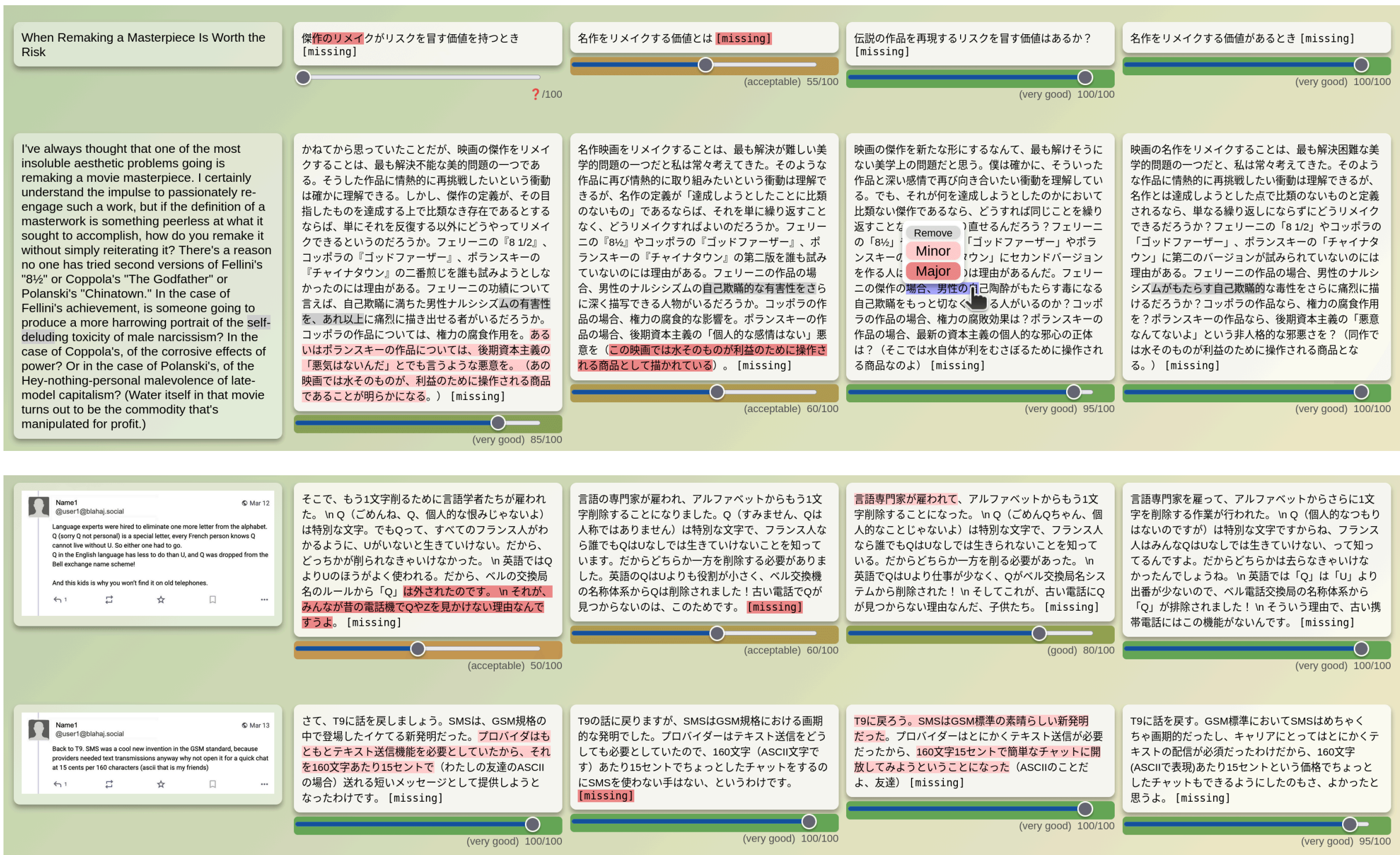

Figure 2: Two screenshots of document-level contrastive ESA (cESA) with four model outputs shown next to each other. Each segment translation has a marked list of error spans minor or major and final score from 0 to 100%. See more screenshots and annotation guidelines in Appendix A and Appendix B.

MQM, which yielded higher-quality annotations. However, this approach does not clearly scale up beyond two document outputs. Increasingly, automated evaluation metrics have also been transitioning from single input that is being scored to taking multiple translations at the same time (Züfle et al., 2025; Moosa et al., 2024).

Outside of machine translation evaluation, the dominant paradigm has been comparison, due to the absence of absolute score evaluation protocols. For example, Chatbot Arena (Chiang et al., 2024) and K-Sort Arena (Li et al., 2025) has the crowd choose between two models and $K$ outputs respectively. EASL and RankME (Sakaguchi and Van Durme, 2018; Novikova et al., 2018) combine ranking and absolute score assessment where, similar to contrastive ESA, multiple outputs are shown next to each other. However, instead of ranking, each output is assigned with a score.

The disadvantage of such contrastive approaches is that either (1) all models are at some point compared to all other models uniformly, or (2) a mechanism needs to be devised to obtain a fair model comparison post-hoc. This is because in the second case, some models might have been compared to only the best models and are thus at a disadvantage. For the second option, Callison-Burch et al. (2007) suggest average rank, but which prevents absolute measurements (i.e. average translation quality). There are many other nuanced approaches, which typically use some parametric Bayesian modelling, such as via Trueskill (Herbrich et al., 2006; Minka et al., 2018), to obtain the model ability (Sakaguchi et al., 2014; Sakaguchi and Van Durme, 2018; Chiang et al., 2024). This comes at the cost of having the final model ranking be latent and given by a parametric model, which makes it difficult to naturally interpret, as opposed to a simple average across evaluation items. Instead, we treat this as a human-focused problem such that the final ranking can be obtained non-parametically.

## 3 The cESA annotation protocol

In this section we define the contrastive error span annotation (cESA) protocol in detail. The cESA protocol shows a sequence of source documents on separate screens (one shown in Figure 2). On each screen, the source document is split across multiple segments and translated $k$ times with different models or translators. The task of annotators is to mark error spans in the translations with minor or major severity and assign a final score from 0% to 100% to each translation. Annotators have

• ***85-100% (Very Good)*** *Complete meaning transfer; perfectly natural; no or minimal proofreading.*
• ***65-80% (Good)****: Near-complete transfer, minor inaccuracies; mostly natural, minor awkwardness; needs light proofreading.*
• ***45-60% (Acceptable)****: Main ideas conveyed, noticeable inaccuracies or omissions; uneven naturalness, awkward phrasing; usable only after substantial revision.*
•***25-40% (Borderline)****: Partial transfer; frequent misinterpretation or omission confusing the message; often unnatural; requires major rewrite.*
• ***0-20% (Not acceptable)****: Violation of meaning; large portions mistranslated, missing, or incoherent; unusable without complete retranslation.*

Listing 2: Score anchors for cESA scores.

the opportunity to navigate to previous screens to update their annotations. In contrast to the existing ESA annotation protocol (Kocmi et al., 2024), the possible scores are defined with steps of 5% (so 0%, 5%, 10%, …, 100%), with more detailed anchors of what specific scores mean and tutorials. These anchors should also be used to objectively interpret cESA-annotated data (Listing 2).

See the detailed annotation guidelines and interactive tutorial, shown to all participants at the beginning, in Appendix A.

## 4 Human Evaluation

We run two human evaluation campaigns, which verify the efficacy of our two changes to the ESA annotation protocol in isolation:

- Small-scale standard ESA vs. cESA ($k = 1$) annotation of four translation models.
- Large-scale cESA annotation of ten translation models with $k \in \{1, 2, 3, 4\}$ number of model outputs shown at once on the same screen.

We re-use the data from the WMT 2025 general translation shared task (Kocmi et al., 2025b), specifically the English→Japanese translation subset with 97 segments across 38 documents in total, balanced across diverse domains of news, video, and social screenshots. These inputs vary in length between 1 up to 4 large segments (paragraphs) per document and between 10 to 120 words per segment.

The small-scale campaign covers 4 translation models with 1 extra duplicated translation to compute the inter- and intra-annotator agreement (i.e., will the same annotator give the same ratings to the same output in the same condition; Abercrombie et al., 2025). The large-scale campaign covers ten translation models[3] with 2 extra duplicated translations to esimate annotator agreement and also to align it such that all configurations (cESA with $k \in \{1, 2, 3, 4\}$) can show all translations across 12, 6, 4, and 3 screens, respectively. One of the translations is a human translation. All evaluations are run twice with different sets of annotators to estimate inter-annotator agreement (measuring the consistency across annotators).

The human experiment is designed with a within-subject design and randomly ordered conditions to eliminate subjective and learning effects on average. The experiment is controlled such that the annotator sees all translations for a particular document split across multiple screens, but only in a particular condition. For example, if one annotator saw a specific document in cESA (k=3), they saw all 12 translations (2 of which are duplicates) across multiple screens. They did not see the same document in other configurations, such as cESA (k=1). The annotators are professional translators fluent in English and Japanese, similar to the annotation crowds used in evaluation campaigns, such as WMT (Kocmi et al., 2025b).

They are shown the overall guidelines (Appendix A) but also an interactive tutorial (Appendix C). All annotations happen in the otherwise same annotation interface. The annotators could also navigate to previously annotated items and update their annotations. After completing the annotations, annotators filled an exit questionnaire.

### 4.1 Evaluation results

We show the overall aggregate results across both evaluation campaigns in Table 1, which we now discuss. Beyond annotation speed, which is a proxy for evaluation costs, we are interested in the following measurements:

- Intra-annotator agreement: how similar are two annotations of the same translations by the same annotator? This is quantified by mean absolute error between the two annotations.
- Inter-annotator agreement: how similar are two annotations of the same translations by two different annotators? This is quantified by mean absolute error between the two annotations.

[3]Gemini-2.5 Pro (Gemini Team, 2025), Algharb (Wang et al., 2025), Command-A-MT (Kocmi et al., 2025a), Claude 4 (Anthropic, 2025), DeepSeek-V3 (DeepSeek-AI, 2025), Online-B (anonymized publicly available MT), Mistral-Medium (Mistral AI, 2025), Yolu (Kocmi et al., 2025b, appendix C.33), Laniqo (Guttmann et al., 2025), and professional human translation. The small-scale campaign has human translation, Gemini, Online-B, and Laniqo translations.

| Campaign | Protocol | Time/seg.↓ | Time/err.↓ | InterAA MAE↓ | IntraAA MAE↓ | Stability↑ | Errors | Score distribution |
|---|---|---|---|---|---|---|---|---|
| (ablation) | ESA | 151.0s ±0.04 | 53.9s ±0.01 | 15.9 ±1.167 | 5.9 ±1.003 | 0.964 ±0.001 | 2.8 | 0 100 |
| | cESA (k=1) | 167.3s ±0.05 | 48.0s ±0.01 | 13.4 ±0.915 | 5.5 ±0.891 | 0.962 ±0.001 | 3.5 | 0 100 |
| (main) | cESA (k=1) | 77.8s ±0.01 | 40.5s ±0.00 | 20.0 ±1.054 | 9.7 ±1.491 | 0.802 ±0.001 | 1.9 | 0 100 |
| | cESA (k=2) | 55.4s ±0.02 | 42.2s ±0.01 | 19.0 ±1.040 | 14.1 ±2.351 | 0.786 ±0.001 | 1.3 | 0 100 |
| | cESA (k=3) | 53.7s ±0.03 | 35.3s ±0.02 | 17.0 ±1.080 | 10.5 ±1.869 | 0.813 ±0.001 | 1.5 | 0 100 |
| | cESA (k=4) | 59.3s ±0.05 | 38.2s ±0.03 | 19.0 ±1.041 | 8.4 ±1.524 | 0.797 ±0.001 | 1.6 | 0 100 |

Table 1: Aggregate results from two human evaluation campaigns (small- and large-scale performed by distinct crowds, separated by —). Time/seg and Time/err are average time per segment (translated "paragraph" in a document) and time per annotated error span. Inter/Intra shows annotator disagreement with themselves and with each other, as measured by mean absolute error (0 is best, 100 is worst). Stability is average model ranking correctness (Kendal $\tau_b$) of a random subset $p \in [10\%, 90\%]$ across 100 runs. Errors list the average number of error spans per segment. The ± shows 99% confidence interval using BCa bootstrapping with 10k samples.

- Stability: if we had only a portion of the data, would we arrive at the same model ranking? This is quantified as ranking similarity (Kendall $\tau_b$) of ranking based on a subset against model ranking on the whole data, averaged across 10%, …, 90% of the data.

**Main result: cESA (k=1) is higher quality than ESA.** The top part of Table 1 shows the results of small-scale ESA vs cESA (k=1). By simply updating the guidelines and the interface (clearer guidelines, detailed tutorial, color-coded discretized scales), we can obtain lower inter- and intra-annotator disagreement and higher stability. This is despite the annotators already having worked with the ESA protocol before but not cESA.

This is confirmed by the annotators specifically preferring the discretized (3.3 vs 3.8 on Likert-5) and anchored (3.5 vs 4.2 on Likert-5) scale.

**Main result: Annotating three models at the same time is faster.** The bottom part of Table 1 shows the results for the large-scale cESA evaluation with k=1, 2, 3, and 4.[4] By increasing the number of shown model outputs next to each other, we lower the average annotation time. However, this lowering happens just once when we transition from k=1 to k=2 model outputs and does not decrease further with more outputs. The optimum, in terms of annotation time and quality (indicated by lowest disagreement and highest stability), is when k=3 model outputs are shown next to each other. Same patterns occur across different input types (news, social media screenshots, videos), as shown in Table 2.

[4]The difference in cESA k=1 in small-scale (top) and cESA k=1 in large-scale (bottom) evaluation is caused by the difference in the annotation crowd and the selection of models.

| Protocol | News | Social | Speech |
|---|---|---|---|
| cESA (k=1) | 81.1s ±0.05 | 55.4s ±0.05 | 100.3s ±0.02 |
| cESA (k=2) | 51.0s ±0.12 | 39.2s ±0.26 | 87.0s ±0.05 |
| cESA (k=3) | 43.0s ±0.16 | 55.7s ±0.36 | 71.3s ±0.07 |
| cESA (k=4) | 57.5s ±0.31 | 42.7s ±0.39 | 85.2s ±0.16 |

Table 2: Average time per segment for large-scale campaign (Table 2 second column, bottom part) split across different input types. *News* are text-only, *Social* contain screenshots of social media posts, and *Speech* are short narrated videos.

Again, the preference is corroborated by the annotators themselves stating their maximum number of models they can annotate at the same time (2, 2, 3, 3, 3, 4).

## 4.2 Potential Biases

We now focus on cESA only and address concerns of potential biases in presenting multiple outputs at the same time. While we find some evidence of bias on individual ratings as expected, their practical impact on quality evaluation remains negligible.

**Position bias.** The translation's position on the screen could potentially skew its score, because, for instance, the first shown translation could be inspected with more scrutiny. This is a common confounder occurring in most crowdsourcing campaigns (Lee et al., 2019). We mitigate the impact on final model ranking (Table 3) by shuffling the order, though we still investigate the severity of this bias. Contrary to expectations, the results shown in Table 4 show that there is very little distributional differences based on the position on the annotation screen.

| | cESA (k=1) | cESA (k=2) | cESA (k=3) | cESA (k=4) |
|---|---|---|---|---|
| Gemini-2.5-Pro | 80.4 | 85.4 | 83.2 | 82.5 |
| Algharb | 80.1 | 86.2 | 82.6 | 82.4 |
| Human | 75.5 | 81.2 | 81.3 | 81.1 |
| CommandA-MT | 76.5 | 82.2 | 78.1 | 78.2 |
| Claude-4 | 76.1 | 81.2 | 78.9 | 78.4 |
| DeepSeek-V3 | 72.5 | 80.7 | 77.1 | 77.5 |
| ONLINE-B | 72.4 | 78.6 | 76.0 | 75.0 |
| Mistral-Medium | 70.9 | 77.7 | 76.1 | 76.3 |
| Yolu | 66.4 | 75.0 | 70.8 | 70.3 |
| Laniqo | 58.0 | 68.7 | 64.5 | 66.0 |

Table 3: Model ranking according to the four configurations of cESA (main campaign). The final model ranking is based on collected score averages.

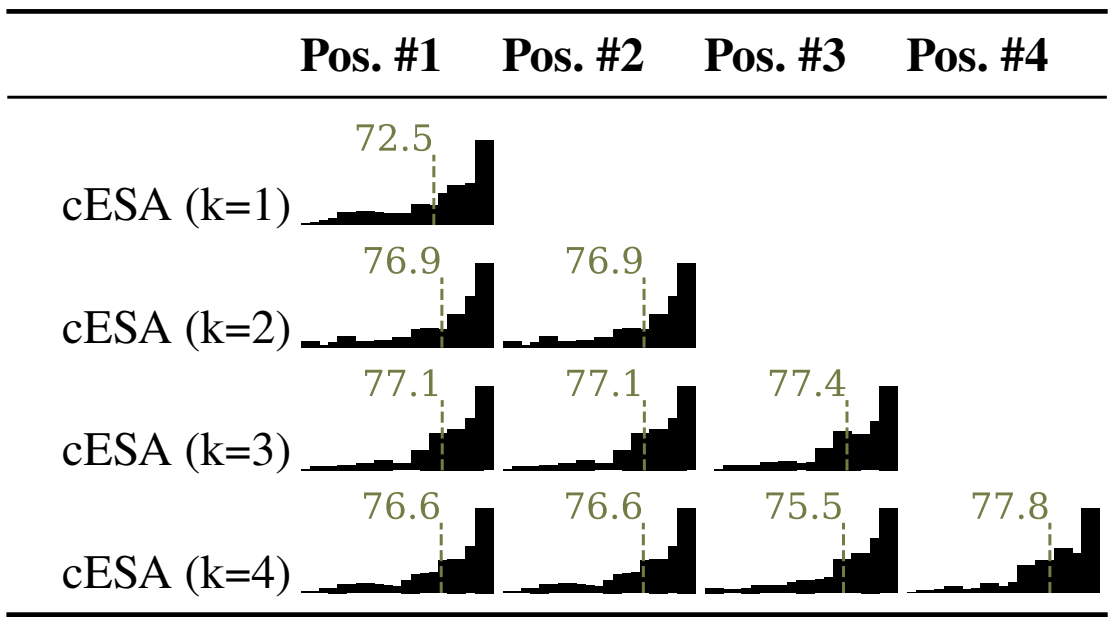

Table 4: Score distribution of translations shown at a particular position on each screen (main campaign).

| | Max sim | Mean sim |
|---|---|---|
| cESA (k=2) | 0.4 | 0.4 |
| cESA (k=3) | 0.0 | −0.4 |
| cESA (k=4) | −0.5 | −1.1 |

Table 5: Differences in specific model output's scores in cases it was in a more similar context than on screens with more dissimilar other model outputs. The contexts are computed either with "maximum similarity" or "mean similarity".

**Surrounding similarity bias effect.** We investigate to what extent translations similarity affects the number of annotated errors. One possible problem is that when similar model outputs are presented, they will receive more annotated errors because the annotator already knows what to look for and can thus spend the rest of the annotation effort on finding more errors.

A direct correlation between similarity and assigned scores is confounded by source difficulty: easy and short inputs naturally result in both high similarity and high scores, consistent with minimum Bayes risk (Müller and Sennrich, 2021). We instead test a counterfactual proxy: if a specific model had less similar neighbors, would it receive fewer error annotations? Our experimental setup permits this measurement, as identical models are evaluated across multiple screens in varying contexts. For example, assume that for a particular translation, model A's output is very similar to B's and C's output but dissimilar to D and E. Then, if similarity does bias the scoring in the way that in a more similar context a given translation gets lower scores, the score of A in the following configuration A B C would be lower than the score of A in A D E.

We compute the similarity using symmetric chrF (Popović, 2015)[5] and consider comparing similarity within screens based on maximum or mean similarity. This similarity is computed between A and the rest of the screen. Each screen with A then gets a single scalar (maximum or mean) similarity based on which we compute the final statistics.

The proportion of times this happens is shown in Table 5, which hints at almost no effect of −0.2 points as sometimes more similarity increases the model score and vice versa.

**Surrounding quality bias effect.** We also analyze the effect of surrounding quality to determine whether the presence of strong models unfairly penalizes weaker ones, or whether the presence of weak models artificially inflates the reported performance of stronger models. This is occasionally referred to as the decoy effect (Eickhoff, 2018) and has been observed in translation evaluation (Bojar et al., 2011, 2016; Graham et al., 2014). We use a similar strategy as for similarity bias. If C is much better than A and B, would evaluation in A C lead to a lower score of A than evaluation in A B ?

The effect size is shown in Table 6, which hints at a very small but consistent bias: the presence of a stronger model on the screen causes model output to receive, on average −0.5 lower score.

[5]We chose a surface-level string similarity intentionally because a semantic-based one would abstract away from translation model paraphrases and erroneous choices.

| | Max score | Mean score |
|---|---|---|
| cESA (k=2) | −0.3 | −0.3 |
| cESA (k=3) | −0.6 | −1.4 |
| cESA (k=4) | −0.2 | −0.7 |

Table 6: Differences in specific model output's scores in cases it was in a more "competitive" context than on screens with lower-scoring other model outputs. The contexts are computed either with "maximum score" or "mean score".

| | Within-screen | Between-screen |
|---|---|---|
| cESA (k=1) | N/A | 22.2 |
| cESA (k=2) | 14.7 | 18.9 |
| cESA (k=3) | 12.9 | 17.8 |
| cESA (k=4) | 15.0 | 19.6 |

Table 7: Average absolute score difference between model outputs on the same screen (within-screen) and across screens (between-screen).

This is less than model average differences (see Table 3) and also alleviated by the model outputs per screen chosen randomly.

**Within-screen calibration.** Lastly, we discuss the calibration effect of multiple model outputs on the same screen. The overall distribution and span of the score is the same across all configurations cESA (k=1, 2, 3, 4); see last column of Table 1. At the same time, the difference between the scores of the same model output on the same screen and even across screens is smaller, as per Table 7. This means that simply having multiple models next to each other helps with higher annotation consistency.

## 5 Conclusion

In this work we advanced the state-of-the-art human annotation protocol for translation evaluation by building on existing ESA and introducing its updated and contrastive version, cESA. We evaluated it on diverse inputs of varying modalities and lengths and showed that it consistently yields higher-quality annotations than available alternatives. At the same time, we found no impactful biases that are introduced by this setup and the collected annotation scores can be used directly as averages for model comparison.

For evaluation practitioners, we recommend showing k=3 model outputs next to each other, which balances the speed up in time with annotation quality and does not overstrain annotators.

For ease of use, this annotation protocol is implemented in an existing ready-to-use annotation platform (Listing 1) together with tutorials in 18 language pairs.

## Limitations

There are other potential biases in human annotation protocols, such as the order in which model outputs are presented (Knowles, 2021). We sidestep this by shuffling the documents and by allowing annotators to go back and change previous responses, which was not the case in previous annotation interfaces.

## Ethics statement

The annotators were paid a standard industry wage and gave consent for their anonymized data to be used in a scientific study and published. The data does not contain any potentially harmful content.

## Acknowledgements

Vilém Zouhar gratefully acknowledges the support of the Google PhD Fellowship.

## A cESA and ESA Annotation Guidelines

The full annotation guidelines are shown in Listing 3 and Listing 4. They are included by default when using the Pearmut tool (Listing 1). The annotation guidelines are shown at the top of the screen at all times (Figure 3).

***Task****: Read the source text and competing translations. Highlight all errors. Rate each translation (each column is an output from a single model/translator). At the end, click Next in top-right corner.*

***Highlighting errors:***

- *Select text containing an error to mark it as Minor. Then assign severity by hovering over the error span.*
- *Minor error: Imperfections or stylistic issues that do not impact the core message (e.g., awkward phrasing).*
- *Major error: Confuses meaning, misrepresents the source, or violates the message (e.g., incorrect information, confusing wording).*
- *Missing text: Highlight the* `[MISSING]` *tag if the translation omits important source text.*

***Important rules:***

- ***Multiple errors:*** *Use separate highlights for each error.*
- ***Hallucinations:*** *Highlight unsupported extra text; mark as Major.*
- ***Wrong language:*** *Highlight the entire text, mark as Major, and assign a score of 0.*
- ***Consistency:*** *Check translation consistency (e.g., technical terms) across the document.*

***Rating scale (0-100%):***

- *85-100% (Very Good): Complete meaning transfer; perfectly natural; no or minimal proofreading.*
- *65-80% (Good): Near-complete transfer, minor inaccuracies; mostly natural, minor awkwardness; needs light proofreading.*
- *45-60% (Acceptable): Main ideas conveyed, noticeable inaccuracies or omissions; uneven naturalness, awkward phrasing; usable only after substantial revision.*
- *25-40% (Borderline): Partial transfer; frequent misinterpretation or omission confusing the message; often unnatural; requires major rewrite.*
- *0-20% (Not acceptable): Violation of meaning; large portions mistranslated, missing, or incoherent; unusable without complete retranslation.*

*For unexpected problems and remarks, use comment box in ⚙ (top right).*

Listing 3: Annotation guidelines for the cESA protocol (ours).

***Error spans:*** *Click on the start of an error, then click on the end to mark an error span. Hover over an existing highlight to change error severity (minor/major) or remove it.*

***Error severity:***

- ***Minor****: Style, grammar, or word choice could be better.*
- ***Major****: Meaning is significantly changed or is hard to understand.*
- ***Tip****: Mark the general area of the error (doesn't need to be exact). Use separate highlights for different errors. Use* `[missing]` *at the end of a sentence for omitted content. Score each translation using the slider based on meaning preservation and quality.*
- ***Important****: The relative order of scores matters; ensure better translations have higher scores than worse ones.*
  - *0: Broken/Nonsense*
  - *33%: Flawed: substantial issues.*
  - *66%: Good: small issues with grammar, fluency, or consistency.*
  - *100%: Perfect: meaning and style align completely with the source.*

Listing 4: Annotation guidelines for the ESA protocol (previous work).

## B Screenshots

See Figure 3 for screenshots of the four configurations across domains (you will need to zoom in).

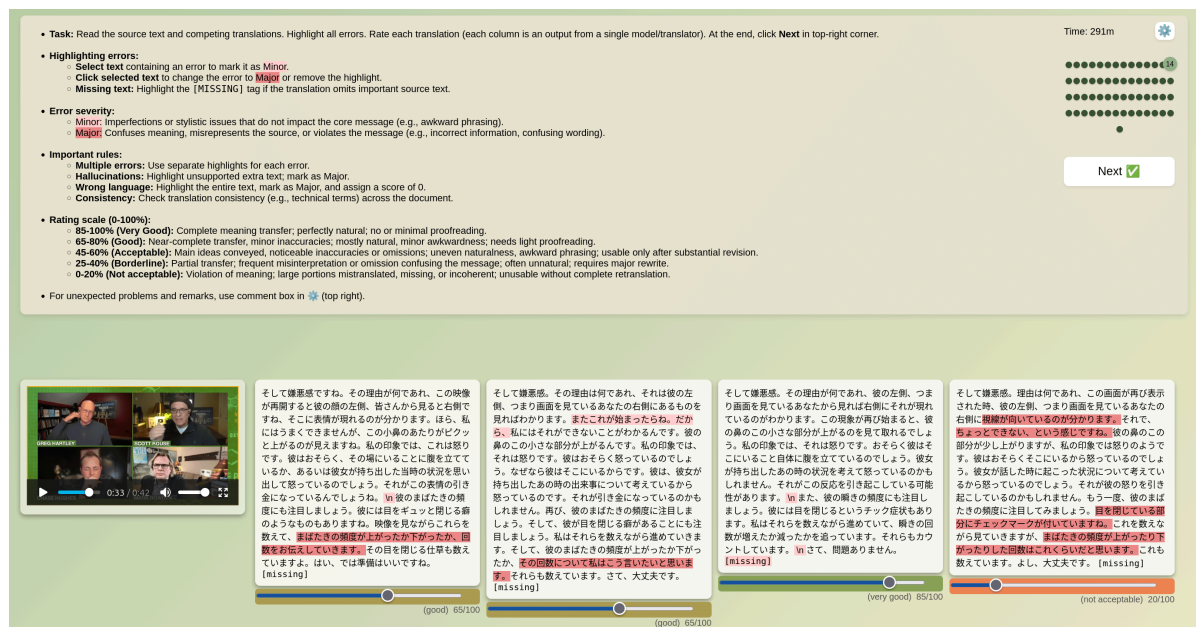

cESA (k=4) with four model translations of a video next to each other and the annotation guidelines at the top.

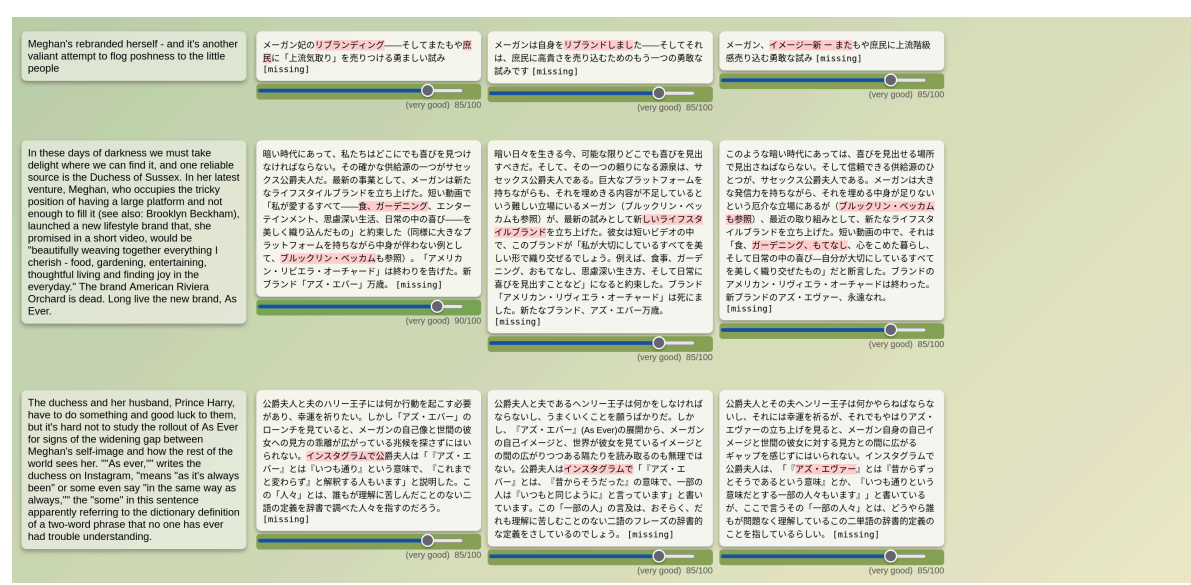

cESA (k=3) with three model translations of a multiple paragraphs in a document next to each other.

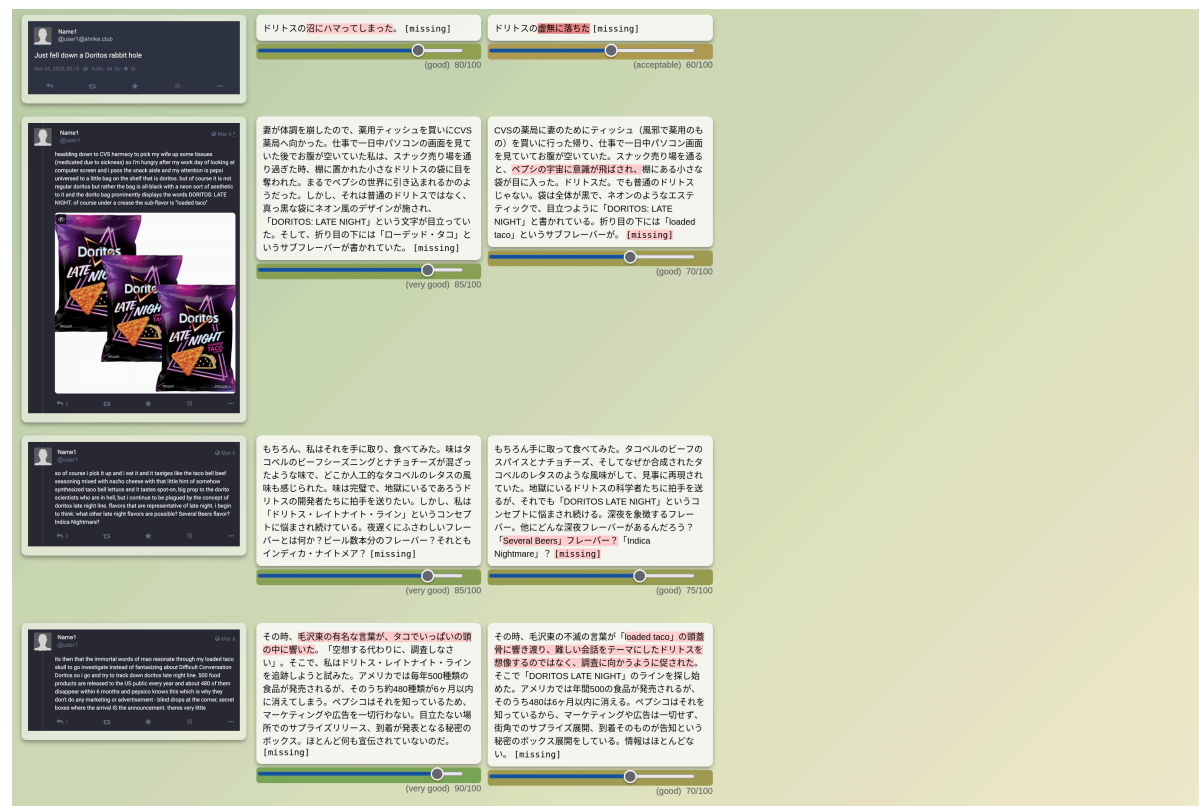

cESA (k=2) with two model translations of a social media thread next to each other.

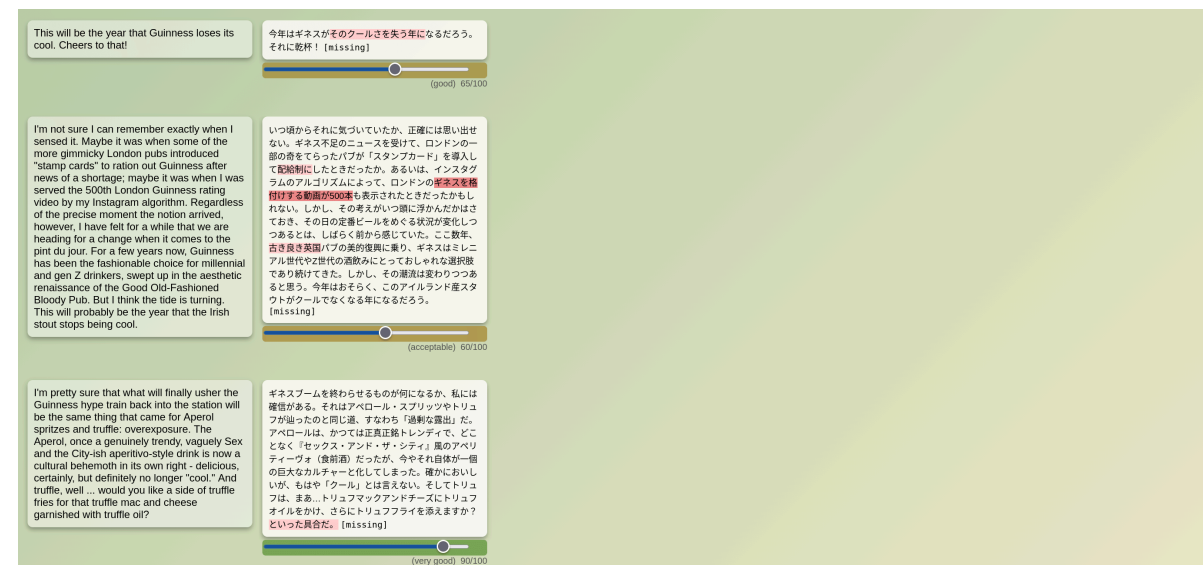

cESA (k=1) with one model translation of a document.

Figure 3: Screenshots of four configurations of contrastive ESA in Pearmut. The widths of the model output panels were intentionally kept the same to avoid any UI-based effects on the annotation.

# C Tutorials

Each annotator was shown a tutorial task in addition to the general guidelines, which they could refer to. These tutorials, shown in Figure 4, are in the format of X→English despite the task being English →X so that we guarantee the existence and severity of errors on the English side. We provide these tutorials for future practitioners for Egyptian Arabic, Belarusian, Czech, German, Estonian, Armenian, Indonesian, Icelandic, Italian, Japanese, Kazakh, Korean, Russian, Northern Sami, Thai, Ukrainian, Traditional Chinese, and Chinese into English and can further be extended.

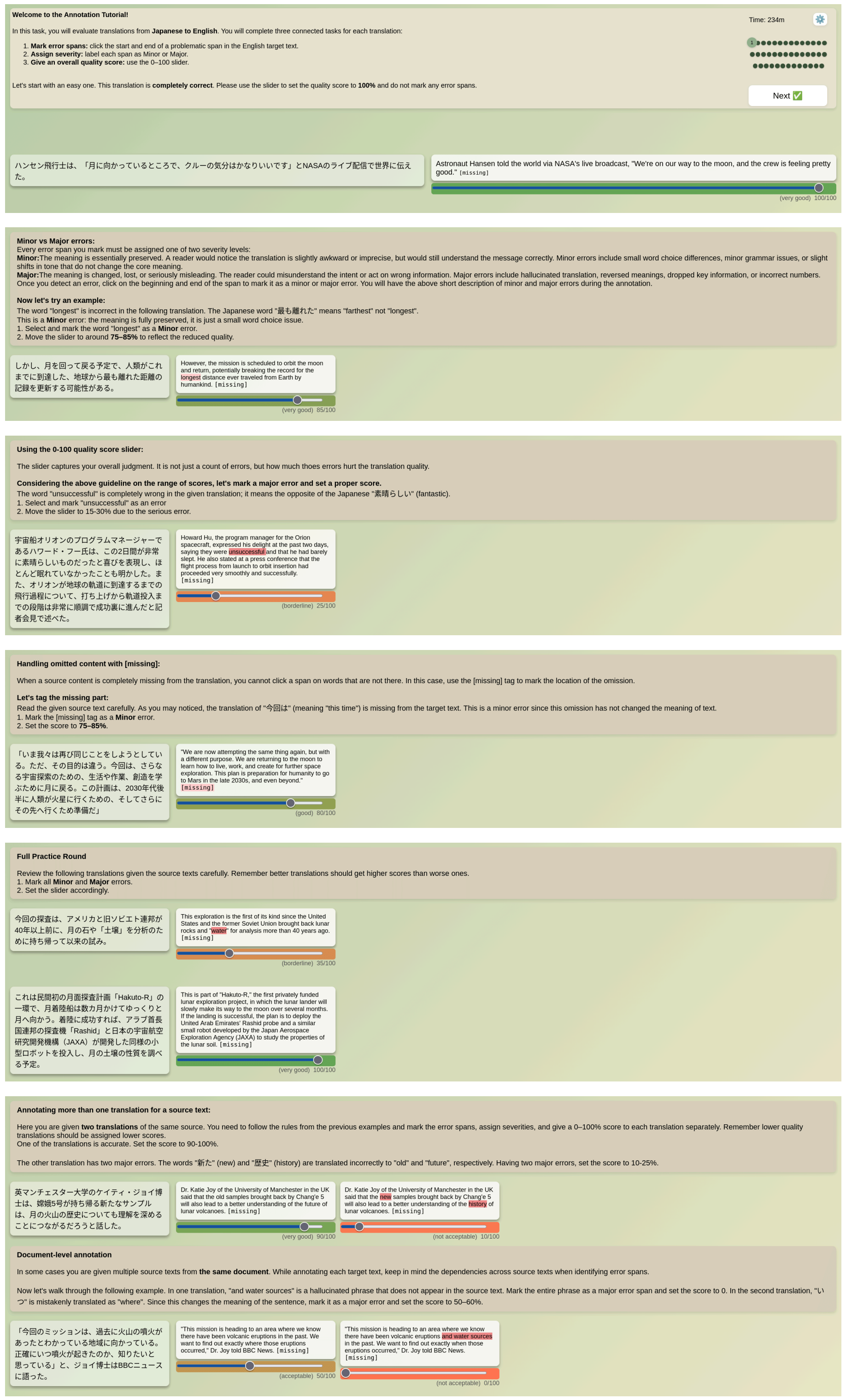

Figure 4: Sequence of six introductory tutorial steps that teaches the annotators the translation evaluation task.